\documentclass{article}
\usepackage{ijcai26}

\usepackage{times}
\usepackage{soul}
\usepackage{url}
\usepackage[hidelinks]{hyperref}
\usepackage[utf8]{inputenc}
\usepackage[small]{caption}
\usepackage{graphicx}
\usepackage{amsmath}
\usepackage{amsthm}
\usepackage{booktabs}
\usepackage{algorithm}
\usepackage{algorithmic}
\usepackage[switch]{lineno}
\usepackage{array}
\usepackage[margin=1in]{geometry}
\usepackage{booktabs}
\usepackage{array}
\usepackage{makecell}
\usepackage{xcolor}
\usepackage{multirow}

\definecolor{thrblue}{RGB}{0,76,204}
\definecolor{sporange}{RGB}{230,120,20}

\newcommand{\scell}[3]{%
\begin{tabular}[c]{@{}c@{}}
#1\\
{\color{blue}#2}\\
{\color{orange}(#3)}
\end{tabular}}

\definecolor{thrblue}{RGB}{0,76,204}
\definecolor{sporange}{RGB}{230,120,20}

\def\placeholder{Window-Diffusion}
\title{\placeholder: Accelerating Diffusion Language Model Inference with Windowed Token Pruning and Caching}

\author{
Fengrui Zuo$^{1*}$
\and
Zhiwei Ke$^{1}$\thanks{Equal contribution.}
\and
Yiming Liu$^{1}$
\and
Wenqi Lou$^{1,2}$\thanks{Corresponding author.}
\and
Chao Wang$^{1\dagger}$
\and
Xuehai Zhou$^{1}$\\
\affiliations
$^{1}$School of Computer Science and Technology, University of Science and Technology of China\\
$^{2}$Suzhou Institute of Advanced Research, University of Science and Technology of China\\
\emails
\{frzuo,kzw17613940379,yimingliu\}@mail.ustc.edu.cn,
\{louwenqi,cswang,xhzhou\}@ustc.edu.cn
}

\begin{document}

\maketitle

\begin{abstract}

Diffusion language models (DLMs) generate text through iterative denoising, but inference requires full-sequence attention at every iteration, resulting in substantial redundant computation on masked tokens. Block-wise diffusion can reduce this cost, yet it typically relies on retraining and constrained update orders, limiting its direct applicability to pretrained DLMs.
Our token-level analysis reveals pronounced structural locality in DLM inference. 
Decoding is driven by a small set of prefix-localized active tokens; the influence of distant undecoded context diminishes rapidly, and decoded tokens exhibit stage-wise temporal stability, enabling reuse of intermediate representations except for a brief post-decode transient.
Motivated by these observations, we propose \textbf{\placeholder}\footnote{The source code is available at https://github.com/vhicrgit/Window-Diffusion.}, a window-based token pruning and caching method for inference. We maintain a local computation window that slides rightward as denoising progresses, and partition undecoded tokens into: (i) \textit{active tokens} that are computed online, (ii) \textit{buffer tokens} whose KV states are cached and periodically refreshed, and (iii) \textit{far-field tokens} that are pruned outside the window. Computation is restricted to active and buffer tokens within the window, while far-field tokens are omitted at each stage.
Experiments on LLaDA and Dream show that, under matched compute budgets, our method achieves up to $99\times$ inference speedup while largely preserving generation performance.

\end{abstract}
\section{Introduction}
Diffusion Language Models (DLMs) have recently gained attention as an alternative to autoregressive generation \cite{li2025aso}. Unlike Large Language Model (LLM) that generate tokens left-to-right, DLMs cast text generation as iterative denoising over the full sequence: inference starts with the prompt plus many mask tokens and progressively refines them over multiple iterations into coherent text. This sequence-level paradigm enables parallel decoding and global modeling, allowing DLMs to achieve competitive performance in language modeling, code generation, and reasoning \cite{nie2025largeld,ye2025dream7d,zhu2025llada1v}.

Despite these advantages, DLM inference incurs substantial latency \cite{peng2025howea,zhang2025diffusion,feng2025theoreticalba}. A key bottleneck is that each denoising step performs bidirectional attention over the entire sequence up to a fixed maximum length.
Some prior work improves DLM inference efficiency through model quantization \cite{xu2025dllmquantqd,zhang2025quantdllmpe,lin2025quantizationmd} and KV caching \cite{ma2025dkvcachetc,liu2025dllmcachead}, yielding measurable speedups. However, quantization reduces arithmetic and communication overhead but does not shorten the computed sequence. KV caching reuses the key/value states of decoded tokens to avoid recomputation, but it likewise cannot eliminate the redundant attention and feed-forward cost caused by large numbers of mask tokens.
Recent approaches have sought to reduce per-step cost by restructuring generation (e.g., block diffusion\cite{arriola2025blockdi}, which combines inter-block autoregression with intra-block diffusion to limit computed tokens). However, they often require costly retraining and require block-by-block decoding, thereby sacrificing diffusion’s flexible update order. Consequently, eliminating per-step redundancy from full-sequence computation on pretrained DLMs without retraining remains a key challenge.


To address this challenge, we revisit DLM inference from a token-level perspective and conduct a systematic analysis of decoding behavior and representation dynamics. Our analysis reveals that diffusion-based inference exhibits strong structural locality across both spatial and temporal dimensions. More concretely, we observe that: (i) tokens likely to be updated consistently concentrate near the prefix of the undecoded region, dominating the decoding process;  (ii) the effective dependency of active tokens on undecoded context is limited, with contributions from distant undecoded tokens rapidly diminishing, while non-active tokens in the retained local context exhibit stable intermediate representations that admit caching; and  (iii) decoded tokens display stage-wise temporal stability, where tokens decoded long ago remain highly stable across adjacent diffusion steps, whereas newly decoded tokens undergo a brief post-decode transient before stabilizing.
Together, these observations indicate that full-sequence recomputation at every denoising step is unnecessary, and that DLM inference can be substantially accelerated through token-level selective computation and reuse of cached intermediate states.

Based on these insights, we propose \placeholder, a training-free windowed token pruning and caching strategy for inference in standard DLMs, requiring no architectural or training modifications. Window-Diffusion adopts a dual-window mechanism at each diffusion step, pruning redundant tokens and separating tokens that require accurate online updates from those that primarily serve as context. For the latter, we employ cyclic KV caching with periodic refresh, reusing their key/value states across consecutive steps to avoid repeated low-yield attention computation, while intermittently refreshing to bound approximation error.

Experimental results show that Window-Diffusion markedly accelerates DLM inference across diverse tasks with negligible impact on generation quality, delivering up to 2.3$\times$–6.6$\times$ speedup over the original model and consistently outperforming prior acceleration baselines. Furthermore, when combined with adaptive-length inference, our approach can achieve up to $99\times$ speedup by eliminating redundant decoding steps. Moreover, as it is orthogonal to other model acceleration techniques, Window-Diffusion offers strong potential for further efficiency gains when combined with complementary methods.
\section{Related Work}
\paragraph{Diffusion Language Models.}
Diffusion models were first developed for continuous domains and later extended to discrete sequence modeling. D3PM introduced a general discrete diffusion framework for categorical data \cite{Austin2021StructuredDD}, followed by SEDD with a more stable score-entropy objective \cite{lou2023discretedm}. MDLM showed that masked diffusion can match autoregressive language modeling performance \cite{sahoo2024simpleae}, and subsequent large-scale studies confirmed its favorable scaling on text \cite{nie2024scalingum}. More recently, LLaDA and Dream extended masked diffusion to billion-parameter language models and achieved competitive results on diverse tasks \cite{nie2025largeld,ye2025dream7d}.

\paragraph{Block Diffusion Paradigm.}
To reduce the high inference cost of standard diffusion models, Block Diffusion adopts an autoregressive-over-blocks and diffusion-within-blocks paradigm, enabling variable-length generation while preserving parallel refinement \cite{arriola2025blockdi}. Building on this, Fast-dLLM, D2F, and SDLM further improve efficiency via KV reuse, distillation, and unified next-token/next-block prediction \cite{wu2025fastdllmta,wang2025diffusionlc,liu2025sequentialdl}. However, these methods typically require retraining; moreover, the block-wise decoding imposed by the block diffusion paradigm partially compromises the flexible update order inherent to diffusion models.

\paragraph{KV-Cache in Diffusion Language Models.}
Applying KV-Cache to diffusion models is challenging because token representations evolve under bidirectional attention and decoding order is flexible. dKV-Cache addresses this by selectively caching decoded tokens with delayed updates, achieving significant acceleration on 7B-scale models \cite{ma2025dkvcachetc}. Block diffusion methods also indirectly enable KV reuse via block-level causal structures \cite{arriola2025blockdi,wu2025fastdllmta,wang2025diffusionlc,liu2025sequentialdl}. While caching can reduce redundant computation to some extent, it does not shorten the length of the masked token sequence and therefore cannot fundamentally reduce the computational complexity.

\begin{figure}[t]
    \centering
    \includegraphics[width=\linewidth]{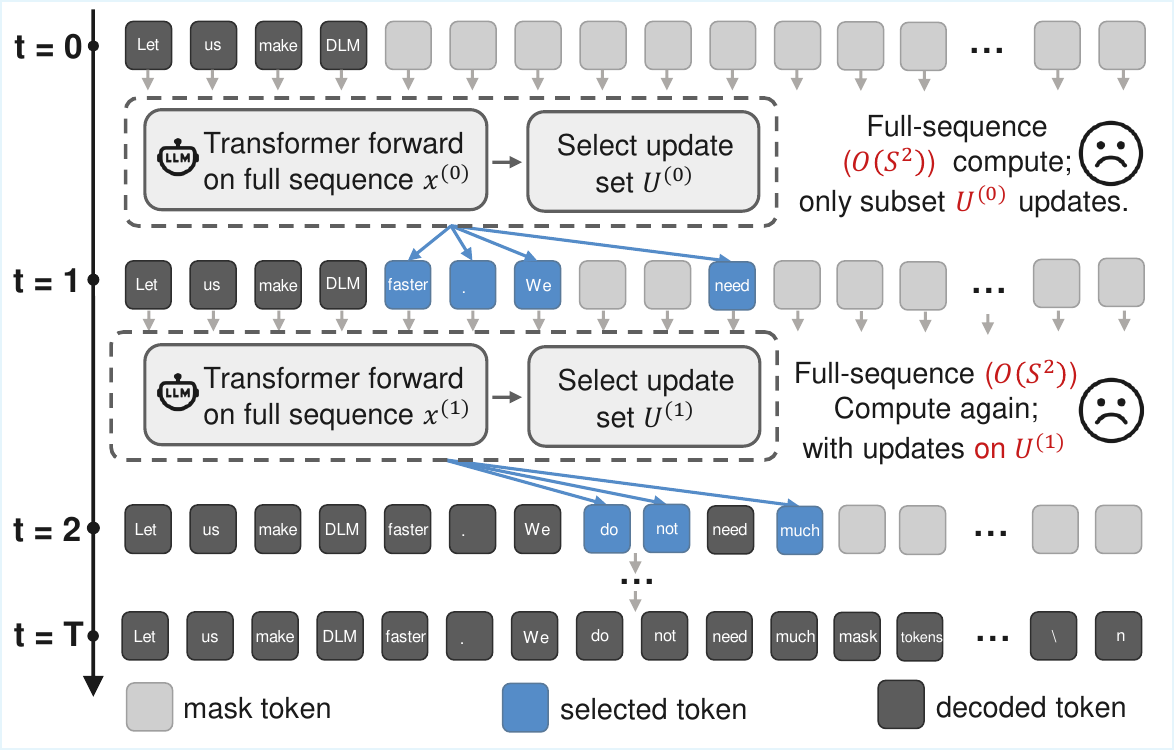}
    \caption{Illustration of mask-based DLM inference.}
    \label{fig:dlm}
\end{figure}

\section{Efficiency Analysis of  DLM Inference}

\subsection{Preliminary: DLM Inference Paradigm}

DLMs generate discrete text sequences through an \textbf{iterative denoising} process. Unlike autoregressive models that decode tokens sequentially, DLMs start from a highly uncertain initial sequence and \textbf{refine undecided positions in parallel} over multiple diffusion steps, gradually approaching the target text distribution. Mask-based discrete DLMs have demonstrated strong scalability in both generation quality and parallelism, and have become the dominant paradigm in recent DLM research. This work focuses on such models and examines the computational characteristics and efficiency bottlenecks of their inference stage.

During inference, the model maintains a fixed-length sequence state
\[
\mathbf{x}^{(t)} = (x^{(t)}_1, \ldots, x^{(t)}_S), \qquad t = 0, 1, \ldots, T,
\]
which contains both decoded tokens and undecided positions. Inference begins from a highly masked initial state, where all non-prompt positions are set to a special placeholder (e.g., [MASK]), and progressively refines the sequence over $T$ diffusion steps.

As illustrated in Fig.~\ref{fig:dlm}, each diffusion step performs a Transformer forward pass over the \emph{entire sequence}. However, only a small subset of positions is actually updated at each step, while the remaining tokens remain unchanged and continue to serve as conditioning context. Let $\mathcal{U}^{(t)} \subseteq \{1,\ldots,S\}$ denote the set of updated positions at step $t$, which typically satisfies $|\mathcal{U}^{(t)}| \ll S$. Despite this sparsity, full-sequence computation is still executed at every step.

From a computational perspective, let $L$ denote the number of Transformer layers. Without causal masking, the dominant cost at each diffusion step arises from full self-attention, yielding an overall inference complexity of
\[
\mathcal{O}(T \cdot L \cdot S^2).
\]
Crucially, although computation scales with the full sequence length, only a small fraction of tokens undergo meaningful state changes at each step. This mismatch between full-sequence computation and sparse token updates represents a major source of redundancy in DLM inference, and directly motivates token-level selective computation and reuse mechanisms explored in this work.

\subsection{Not All Tokens Need to Be Recomputed}
\begin{figure}[thb]
    \centering
    \includegraphics[width=0.99\linewidth]{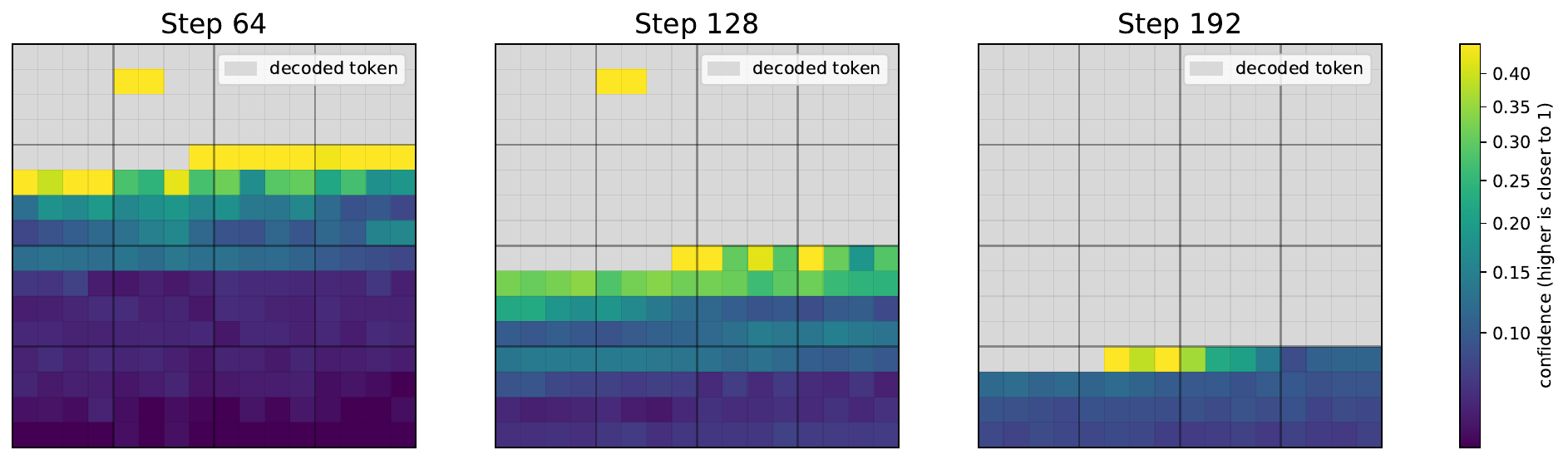}
    \caption{
    Token-wise prediction confidence during diffusion inference.
    The heatmaps visualize prediction confidence over undecoded positions at diffusion steps $t=64, 128,$ and $192$ on \textbf{LLaDA}.
    The total sequence length is 256. Decoded tokens are masked out, and color indicates confidence.
    }
    \label{fig:obs1}
\end{figure}

\begin{figure}[t]
    \centering
    \includegraphics[width=0.9\linewidth]{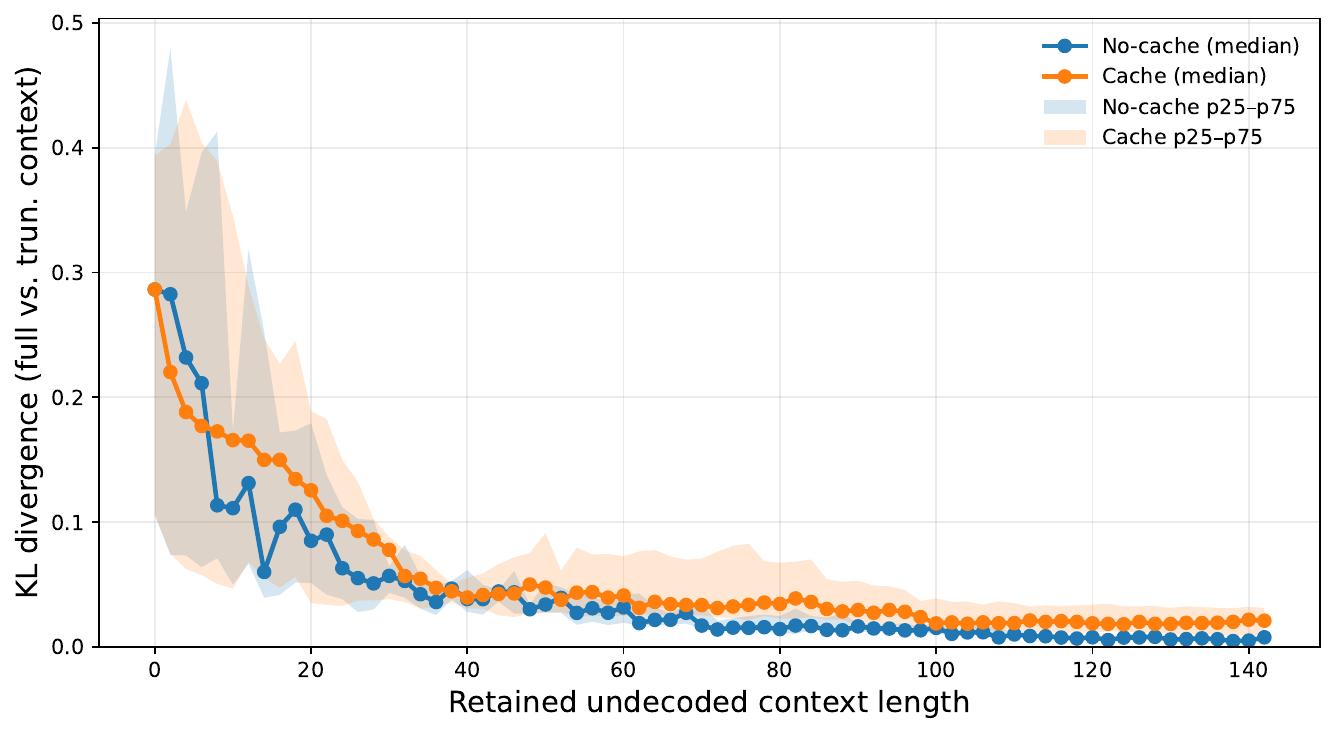}
    \caption{
    KL divergence between active-token predictions under truncated undecoded context and the full-sequence, no-cache reference on \textbf{LLaDA}.
    Results are shown for standard recomputation (\emph{No-cache}) and for reusing KV representations of non-active undecoded tokens from the previous step (\emph{Cache}).
    KL is averaged over active tokens, with shaded bands indicating the 25th--75th percentiles across observation steps (30--60).
    }
    \label{fig:obs2}
\end{figure}

\begin{figure}[t]
    \centering
    \begin{minipage}[t]{0.48\linewidth}
        \centering
        \includegraphics[width=\linewidth]{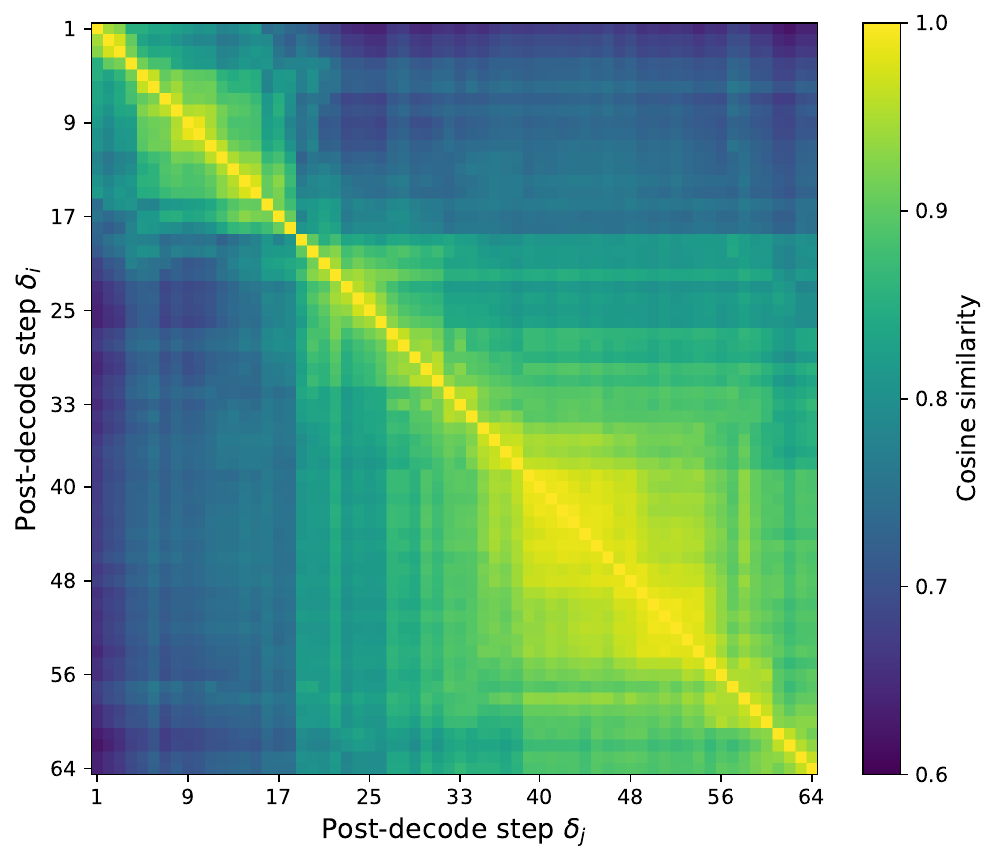}\\
        {\small (a) Recently decoded V similarity}
    \end{minipage}
    \hfill
    \begin{minipage}[t]{0.48\linewidth}
        \centering
        \includegraphics[width=\linewidth]{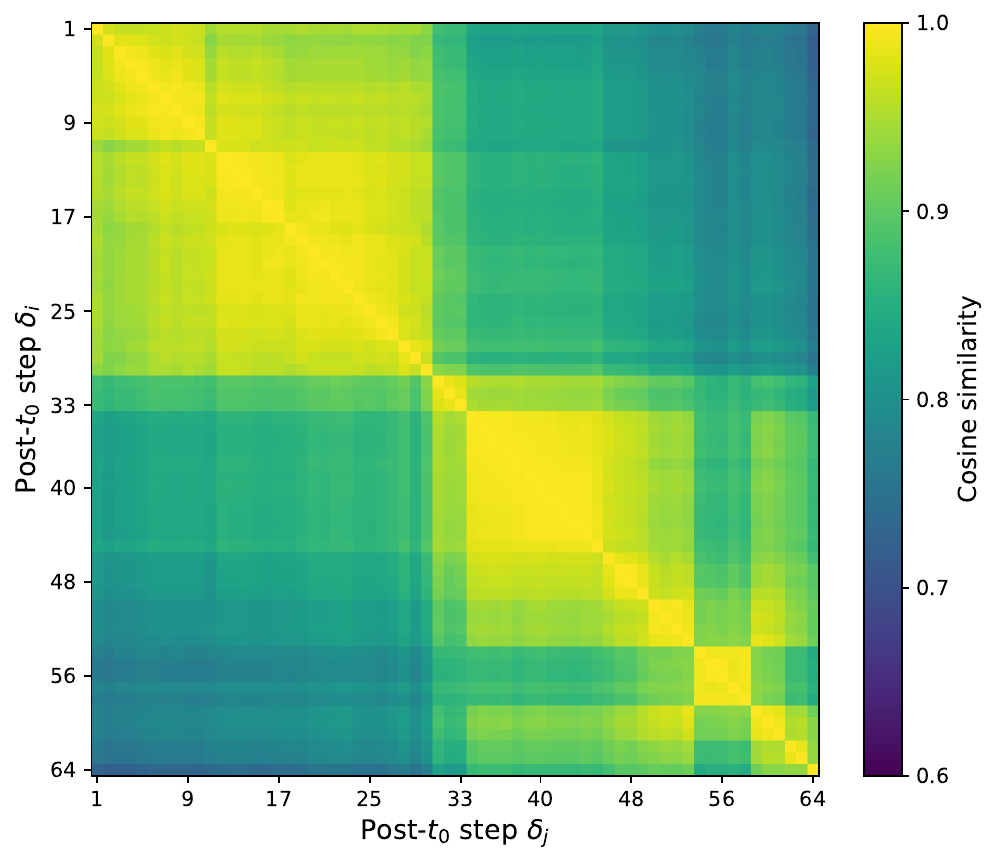}\\
        {\small (b) Earlier-decoded V similarity}
    \end{minipage}
    \caption{
Temporal stability of decoded-token \emph{Value} representations on \textbf{LLaDA}.
(a) Recently decoded tokens: we track tokens at positions 100--120 and compute the average cosine similarity of their \emph{Value} representations over the first 64 steps after each token is decoded.
(b) Earlier-decoded tokens: at an observation step $t_0=64$, we select the first 16 already decoded tokens (excluding the prompt) and measure the average \emph{Value} similarity across the subsequent 64 diffusion steps.
}

    \label{fig:obs3}
\end{figure}


We conduct a systematic token-level analysis of the DLM inference process on the MBPP benchmark to assess whether redundant recomputation can be avoided and to distill empirical insights for inference acceleration. For clarity, we partition tokens into two categories: \textit{active tokens}, which are most likely to be decoded at the current step and therefore require prediction; and \textit{context tokens}, whose primary role is to provide conditioning information to the model. Under this formulation, we examine computational redundancy from three complementary analytical perspectives: (1) the spatial distribution of active tokens along the sequence; (2) the effective dependency range of active tokens on undecoded context; and (3) the temporal evolution of representations for decoded tokens. Empirical analyses along these dimensions reveal several stable and structurally consistent behaviors in DLM inference, which hold across two representative models (LLaDA and Dream) and under diverse inference settings.




\noindent\textbf{Observation 1.} \textit{Active tokens exhibit pronounced prefix locality.}

At each diffusion step, we identify the tokens selected into the update set $\mathcal{U}^{(t)}$, which correspond to the \textit{active tokens} defined earlier. Concretely, we compute prediction confidence for all undecoded positions and visualize its distribution along the sequence, while masking out positions that have already been decoded. As shown in Fig.~\ref{fig:obs1}, as diffusion progresses (e.g., $t=64,128,192$), high-confidence tokens are not uniformly distributed across the undecoded region; instead, they consistently concentrate near its prefix. This observation indicates that, although masked DLMs theoretically permit parallel updates at arbitrary positions, decoding in practice is predominantly driven by prefix tokens, while more distant positions are rarely selected for update. Consequently, recomputing these distant tokens at every diffusion step incurs substantial redundant computation.



\noindent\textbf{Observation 2.} \textit{Active tokens exhibit rapidly saturating gains from undecoded context; distant tokens contribute marginally, while the remaining context admits effective caching.}

Prefix locality alone is insufficient to justify token-level pruning. A more fundamental question is whether predicting \textit{active tokens} truly requires conditioning on all undecoded tokens, and whether these context tokens must be fully recomputed at every diffusion step. In other words, beyond assessing whether distant undecoded tokens can be directly removed, we further seek to evaluate whether the non-active portion of the retained undecoded context admits reusable intermediate representations, such that redundant computation can be avoided via caching.

Motivated by this question, we conduct a controlled analysis building on Observation~1. At a chosen observation step $t=t_0$, we select the first 16 undecoded positions as a representative active set, and systematically truncate the visible context: all decoded tokens are always retained, while the undecoded context is restricted to a prefix of length $W$, with $W$ swept over a range of values. For each $W$, we evaluate two configurations, both using the active-token prediction under full-sequence, no-cache inference as the reference: (1) Truncation only, which performs standard inference on the truncated context; and (2) Truncation + Cache, which, under the same truncation, reuses the KV representations of non-active masked tokens in the retained undecoded context from the previous step $t_0\!-\!1$. In both cases, we measure the deviation of active-token predictions from the reference using KL divergence.

As shown in Fig.~\ref{fig:obs2}, increasing $W$ rapidly reduces the KL divergence, which plateaus at relatively small $W$, indicating that distant undecoded tokens provide limited marginal benefit. Moreover, with caching enabled, the active-token predictions remain close to the reference, suggesting that the intermediate representations of non-active masked tokens in the retained undecoded context are highly reusable. Overall, active-token prediction exhibits pronounced diminishing returns with respect to undecoded context: distant undecoded tokens can be pruned, while the non-active portion of the local undecoded context can be further approximated via caching to reduce redundant computation.



\noindent\textbf{Observation 3.} \textit{Recently decoded tokens undergo a brief transient, whereas earlier-decoded tokens remain locally KV-stationary across adjacent steps.}

The first two observations primarily analyze the dynamics of undecoded tokens. We now turn to the behavior of decoded tokens in subsequent diffusion steps, and specifically examine the local stability of their V representations across adjacent time steps, in order to assess their reusability along the temporal dimension.

In our analysis, we distinguish two categories of decoded tokens and study their representational dynamics separately: (i) tokens that have just been decoded, and (ii) tokens that have been decoded for a sufficiently long time. For the former, we track a set of preselected positions and take each token’s decoding step as the temporal origin, measuring the evolution of V-representation similarity over a fixed number of post-decode steps and averaging the results across tokens. For the latter, we fix an observation step $t=t_0$, select the first 16 decoded tokens at that step (excluding the prompt region), and measure the similarity of their V representations over subsequent diffusion steps, again averaging across tokens. Since we empirically observe that V exhibits larger variation than K, we report V-similarity results throughout.

As shown in Fig.~\ref{fig:obs3}, the two categories of decoded tokens exhibit markedly different local temporal behaviors. Tokens that have just been decoded show low V similarity between adjacent diffusion steps immediately after decoding, indicating that their representations are still undergoing rapid adjustment. In contrast, tokens that were decoded earlier maintain high V similarity across adjacent steps over a sustained time window, demonstrating strong local stability. This observation indicates that representational stability of decoded tokens is explicitly time-dependent: tokens decoded long ago can safely reuse their KV representations over short horizons, whereas newly decoded tokens require continued updates during an initial transient phase.



\noindent\textbf{Summary.}
\textit{At the token level, DLM inference exhibits pronounced spatial locality, rapidly diminishing dependence on distant undecoded context, and stage-wise temporal stability in decoded token representations.}

These properties expose substantial token-level redundancy in diffusion-based inference, indicating that full-sequence recomputation at every diffusion step is unnecessary. In practice, decoding is driven by a small set of spatially localized active tokens, while dependence on distant undecoded context rapidly saturates and many intermediate representations—both from local undecoded context and from previously decoded tokens—remain stable across steps. Motivated by these observations, we next introduce an inference framework that combines token-level selective computation with cached intermediate states to substantially reduce the cost of DLM inference.

\section{Method: \placeholder}
\begin{figure*}[t]
    \centering
    \includegraphics[width=\textwidth]{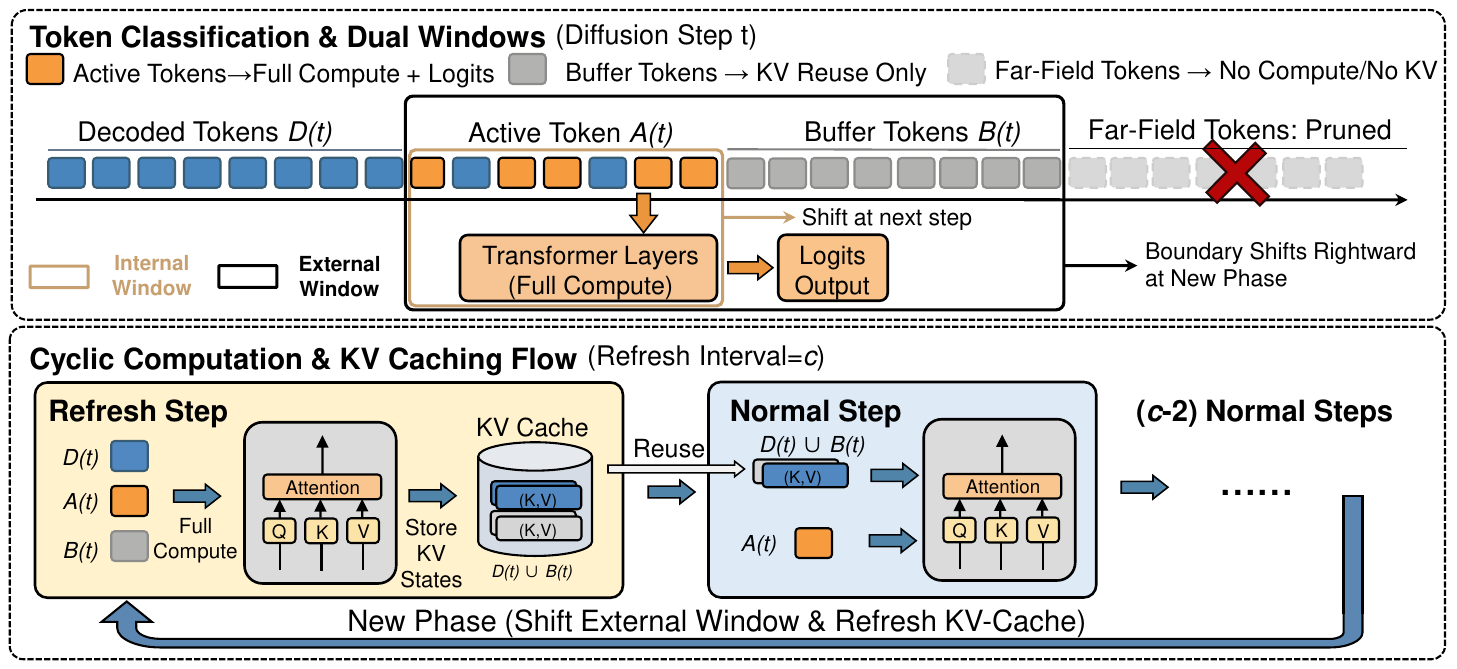}
    \caption{\textbf{Window-Diffusion inference framework.}
    \textbf{Top:} Dual-window token organization within a phase, where only active tokens are fully computed, buffer tokens reuse KV states, and far-field tokens are pruned; the window shifts only at phase boundaries.
    \textbf{Bottom:} Phase-level KV caching, with a refresh step followed by multiple normal steps that reuse cached KV states until the next phase.}
    \label{fig:method}
\end{figure*}
\subsection{Overview}

Based on our preceding analysis of token-level dynamics during inference in DLMs, we propose Window-Diffusion. It is a training-free inference acceleration method that requires no architectural modification, and accelerates masked discrete DLM inference by reducing the number of tokens that participate in critical computations during inference.

Window-Diffusion first partitions the denoising process into a sequence of fixed-length inference phases. Within each phase, it introduces a dual-window mechanism. Specifically, an external window restricts undecoded tokens used as contextual information to a local region near the decoding frontier. On top of this, an internal window further selects a small subset of the most critical undecoded tokens, which form the set of active tokens most likely to be decoded in the current phase. These tokens directly drive decoding progress, such that at each diffusion step prediction distributions (logits) are computed only for these positions. All tokens that have been decoded before the beginning of the current phase always participate in attention as fixed conditional context and are never pruned.

Building on this structure, Window-Diffusion further introduces a phase-level KV caching with periodic refresh mechanism to reduce redundant computation. At the beginning of each phase, a full forward pass is performed over phase-relevant context to initialize reusable Key/Value representations. Within the phase, KV states of most contextual tokens are directly reused, while KV representations of active tokens and tokens newly decoded within the current phase (whose representations are not yet stable) continue to be updated; among them, only active tokens participate in prediction (logit) computation. At phase boundaries, the cache is refreshed to control approximation error.

Overall, Window-Diffusion replaces the conventional inference paradigm of performing full-sequence forward computation at every diffusion step with a structured computation scheme of selective within-phase prediction and stable context reuse, achieving an effective balance between inference efficiency and generation quality. An overview of the framework is shown in Fig.~\ref{fig:method}: the top illustrates dual-window token organization within a phase, while the bottom depicts the phase-level KV caching and refresh process.


\subsection{Dual-Window Inference Mechanism}

As illustrated in Fig.~\ref{fig:method}, Window-Diffusion adopts a dual-window mechanism within each inference phase to organize computation over undecoded tokens. This design preserves the original inference semantics while avoiding repeated full forward passes over all undecoded tokens at every diffusion step. Within a given inference phase $p$, tokens that have already been decoded prior to the phase are always treated as fixed conditional context in attention computation; we denote this set as $\mathcal{D}^{(<p)}$. The dual-window mechanism applies only to tokens that remain undecoded at the beginning of the phase, and constrains their participation and update scope within the phase.

For phase $p$, we define an \textbf{external window} $\mathcal{W}_{\text{ex}}^{(p)}$ as a fixed-length prefix of the currently undecoded region. This window specifies which undecoded tokens participate in attention computation to provide local contextual information. Undecoded tokens outside $\mathcal{W}_{\text{ex}}^{(p)}$ are entirely excluded from computation during the phase and are referred to as \textbf{far-field tokens}. This truncation strategy is motivated by Observation~2, which shows that retaining only the undecoded prefix can closely approximate predictions under full-context conditioning. Within a single inference phase, the position of $\mathcal{W}_{\text{ex}}^{(p)}$ remains fixed; when the phase ends and a new phase begins, the external window is re-determined based on the updated decoded prefix.

Inside the external window, we further define a smaller \textbf{internal window} $\mathcal{W}_{\text{in}}^{(t)} \subseteq \mathcal{W}_{\text{ex}}^{(p)}$, which consists of a small number of tokens at the front of the undecoded prefix at diffusion step $t$. These positions correspond to the tokens that are most likely to be decoded in the current phase, referred to as \textbf{active tokens}, and directly drive decoding progress. We denote their set as $\mathcal{A}^{(t)} := \mathcal{W}_{\text{in}}^{(t)}$. At each diffusion step $t$, prediction distributions (logits) are computed only for the active tokens $\mathcal{A}^{(t)}$ to make decoding decisions.

Let $\mathcal{D}_p^{(t)}$ denote the set of tokens that have been decoded within phase $p$ up to diffusion step $t$. Although these tokens have completed decoding, their internal representations may remain unstable over short time horizons; their computation strategy will therefore be distinguished from that of pre-phase decoded tokens in the subsequent section. The remaining undecoded tokens in the external window, excluding both active tokens and phase-level decoded tokens, form the set of \textbf{buffer tokens}, defined as
\[
\mathcal{B}^{(t)} 
= \mathcal{W}_{\text{ex}}^{(p)} 
\setminus \bigl( \mathcal{A}^{(t)} \cup \mathcal{D}_p^{(t)} \bigr).
\]
As decoding progresses, when some active tokens are successfully decoded and added to $\mathcal{D}_p^{(t)}$, the internal window shifts rightward, promoting new undecoded tokens from the buffer set to maintain a fixed window size and to continue covering the most likely-to-be-decoded positions. Within a phase, buffer tokens typically remain temporarily inactive: they do not participate in decoding decisions, but provide necessary conditional context to active tokens through attention.

Through this dual-window mechanism, the standard DLM inference paradigm—where all undecoded tokens are equivalently computed at every diffusion step—is restructured into a role-based, structured computation scheme. Active tokens are responsible for generating predictions and advancing decoding, buffer tokens and decoded tokens provide conditional context, and far-field tokens are pruned within each phase. This structure also establishes clear operational boundaries for the phase-level KV reuse mechanism introduced next.

\paragraph{Adaptive termination.}
In standard diffusion inference, a maximum generation length must be specified in advance, which may incur unnecessary computation for short outputs. Window-Diffusion naturally alleviates this issue: since the external window always covers only the undecoded prefix, tokens beyond it are automatically pruned. When an end-of-sequence (\texttt{<eos>}) token is generated, the internal window stops advancing, and all subsequent positions remain in the far-field region and are ignored. As a result, generation terminates adaptively based on decoding progress, improving inference efficiency without degrading generation quality.

\begin{table*}[t]
\centering
\footnotesize
\setlength{\tabcolsep}{5pt}
\renewcommand{\arraystretch}{1.15}
\begin{tabular}{p{2.4cm} p{1.4cm} cccc cccc}
\toprule
& & \multicolumn{4}{c}{\textbf{Base}} & \multicolumn{4}{c}{\textbf{Instruct}} \\
\cmidrule(lr){3-6} \cmidrule(lr){7-10}
\textbf{Method} &  &
\textbf{GSM8K} & \textbf{MATH} & \textbf{HumanEval} & \textbf{MBPP} &
\textbf{GSM8K} & \textbf{MATH} & \textbf{HumanEval} & \textbf{MBPP} \\
\midrule

Dream & -- & 77.2 & 39.6 & 57.9 & 56.2 & 81.0 & 39.2 & 55.5 & 58.8 \\

Block Diffusion  & $L = 16$ & 72.3 & 36.4 & 48.1 & 19.6 & 24.3 & 13.3 & 3.0 & 32.2 \\
                 & $L = 32$ & 74.9 & 38.0  & 54.3 & 50.6 & 34.5 & 19.1 & 12.2 & 33.2 \\

Window Diffusion & $L = 16$ & 75.6 & 39.7  & \textbf{55.5} & \textbf{56.8} & 51.2 & 32.2 & 27.4 & 50.8 \\
                 & $L = 32$ & \textbf{76.3} & \textbf{40.3} & 54.3 & 55.2  & \textbf{73.7} & \textbf{33.9} & \textbf{52.4} & \textbf{55.8} \\

\bottomrule
\end{tabular}
\caption{Performance Comparison of Window-Based and Block-Based Token Pruning without KV Caching on Dream Models (L = 16, 32; L indicates window size for Window-Diffusion and block size for Block Diffusion).}
\label{tab:token prunning}
\end{table*}

\begin{table*}[t]
\centering
\small
\setlength{\tabcolsep}{6pt}        
\renewcommand{\arraystretch}{1.15} 

\begin{tabular}{>{\centering\arraybackslash}m{2.0cm} cccc cccc}    
\toprule
& \multicolumn{4}{c}{\textbf{Base}} & \multicolumn{4}{c}{\textbf{Instruct}} \\
\cmidrule(lr){2-5} \cmidrule(lr){6-9}
\textbf{Method} &
\textbf{GSM8K} & \textbf{MATH} & \textbf{HumanEval} & \textbf{MBPP} &
\textbf{GSM8K} & \textbf{MATH} & \textbf{HumanEval} & \textbf{MBPP} \\
\midrule

Dream
& \scell{77.2}{5.3}{1.0$\times$} & \scell{39.6}{4.7}{1.0$\times$} & \scell{57.9}{8.6}{1.0$\times$} & \scell{56.2}{4.7}{1.0$\times$}
& \scell{81.0}{14.4}{1.0$\times$} & \scell{39.2}{8.7}{1.0$\times$} & \scell{55.5}{6.1}{1.0$\times$} & \scell{58.8}{4.7}{1.0$\times$} \\
\midrule

DKV-Cache
& \scell{74.1}{14.8}{2.8$\times$} & \scell{29.5}{10.1}{2.2$\times$} & \scell{\textbf{59.8}}{11.1}{1.3$\times$} & \scell{51.0}{10.1}{2.2$\times$}
& \scell{82.7}{17.4}{1.2$\times$} & \scell{39.0}{11.2}{1.3$\times$} & \scell{33.5}{8.1}{1.3$\times$} & \scell{53.2}{6.4}{1.4$\times$} \\
\midrule

Fast-dLLM
(Prefix-Cache)
& \scell{74.7}{22.5}{4.2$\times$} & \scell{38.0}{16.0}{3.5$\times$} & \scell{54.9}{17.4}{2.0$\times$} & \scell{53.8}{15.9}{3.4$\times$}
& \scell{81.0}{24.9}{1.7$\times$} & \scell{39.2}{17.4}{2.0$\times$} & \scell{54.9}{12.9}{2.1$\times$} & \scell{48.6}{10.0}{2.1$\times$} \\
\midrule

Fast-dLLM
(Dual-Cache)
& \scell{74.4}{24.1}{4.5$\times$} & \scell{37.4}{25.1}{5.3$\times$} & \scell{57.9}{27.5}{3.2$\times$} & \scell{53.2}{25.3}{5.4$\times$}
& \scell{81.4}{26.7}{1.8$\times$} & \scell{\textbf{39.6}}{27.2}{3.1$\times$} & \scell{23.8}{25.9}{4.2$\times$} & \scell{50.2}{24.9}{5.3$\times$} \\
\midrule

\textbf{Window-Diffusion}
& \scell{\textbf{76.1}}{29.9}{5.7$\times$} & \scell{\textbf{39.2}}{29.8}{6.3$\times$} & \scell{57.9}{33.7}{3.9$\times$} & \scell{\textbf{54.4}}{29.7}{6.3$\times$}
& \scell{\textbf{82.9}}{32.5}{2.3$\times$} & \scell{38.5}{33.9}{3.9$\times$} & \scell{\textbf{58.5}}{32.3}{5.3$\times$} & \scell{\textbf{55.4}}{31.1}{6.6$\times$} \\
\bottomrule
\end{tabular}

\caption{Comprehensive benchmark results of various acceleration methods on Dream models over four tasks. Each cell reports accuracy (top row) and decoding throughput in tokens per second with relative speedup to the Dream baseline (bottom row, {\color{thrblue}blue: tokens per second} / {\color{sporange}orange: relative speedup}). Window-Diffusion uses an internal window size of L = 16 with a refresh cycle of 32, and early stopping is disabled. See Appendix \ref{appendix:hyperparameters} for detailed hyperparameters.}
\label{tab:perf_comparison}
\end{table*}

\subsection{KV Caching with Phase-Level Refresh}

Under the dual-window inference structure, each diffusion step only needs to compute prediction distributions (logits) for the active tokens within the internal window to make decoding decisions. Nevertheless, due to attention interactions, active tokens must still attend to contextual tokens from three sources of contextual tokens: buffer tokens within the external window,
tokens decoded prior to the beginning of the current phase, and tokens decoded earlier within the same phase (i.e., $\mathcal{B}^{(t)}$, $\mathcal{D}^{(<p)}$, and $\mathcal{D}_p^{(t)}$). Among these, the first two groups---$\mathcal{B}^{(t)}$ and $\mathcal{D}^{(<p)}$---are typically stable over short time horizons. Therefore, recomputing their attention Key/Value (KV) representations at every diffusion step would introduce substantial redundant computation. To address this inefficiency while preserving inference stability, we further introduce a phase-level KV caching mechanism with periodic refresh.

Concretely, for each inference phase $p$, we partition the diffusion steps into one \textbf{refresh step} followed by multiple \textbf{normal steps}, as illustrated in the bottom part of Fig.~\ref{fig:method}. At the refresh step, denoted by $t_{\mathrm{ref}}^{(p)}$, the model performs a full forward pass over all tokens involved in phase $p$, including the tokens decoded before the phase $\mathcal{D}^{(<p)}$ and the undecoded tokens within the external window $\mathcal{W}_{\text{ex}}^{(p)}$. The resulting Key/Value representations from all attention layers are then written into the cache:
\[
(K,V)^{(t_{\mathrm{ref}}^{(p)})}_{\mathcal{D}^{(<p)} \cup \mathcal{W}_{\text{ex}}^{(p)}}
\;\leftarrow\;
\mathrm{Forward}_\theta\!\left(\mathbf{x}^{(t_{\mathrm{ref}}^{(p)})}\right).
\]

During the remaining diffusion steps within the same phase (i.e., the normal steps), the model reuses the KV cache initialized at the refresh step and applies role-specific computation strategies to different categories of tokens. Specifically, at diffusion step $t$: \textbf{active tokens} $\mathcal{A}^{(t)}$ participate in full forward computation to update their representations and compute prediction distributions (logits), thereby driving decoding progress; tokens decoded within the current phase, $\mathcal{D}_p^{(t)}$, continue to participate in representation updates to refresh their Key/Value states, but no longer contribute to prediction distribution computation; \textbf{buffer tokens} $\mathcal{B}^{(t)}$ act solely as conditional context in attention interactions, and their KV representations are directly reused from the most recent refresh step.
It is important to emphasize that within a single inference phase, tokens that are newly decoded during the phase continue to participate in KV updates for several subsequent diffusion steps, rather than having their representations frozen immediately after decoding. When phase $p$ ends and a new phase begins, the model performs a new KV refresh based on the updated decoding state, and the above procedure is repeated.

Through this phase-level KV reuse with periodic refresh strategy, Window-Diffusion effectively avoids redundant computation over stable contextual tokens across diffusion steps, achieving substantial inference speedups while maintaining stable generation quality.

\section{Experiments}
\subsection{Experimental Setup}

We conduct a systematic evaluation of the proposed Window-Diffusion method on two state-of-the-art diffusion language models, LLaDA and Dream. The experimental data for LLaDA can be found in Appendix \ref{appendix:llada}. Our benchmarks cover four widely used datasets, including GSM8K-CoT~\cite{cobbe2021trainingvt}, MATH~\cite{hendrycks2021measuringmp}, HumanEval~\cite{chen2021evaluatingll}, and MBPP~\cite{austin2021programsw}.

To examine the effectiveness of the pruning strategy, we first compare Window-Diffusion with Block Diffusion \cite{arriola2025blockdi}, a token-pruning-based baseline, to evaluate how different pruning schemes affect performance preservation. We further compare against representative inference acceleration methods to assess the efficiency gains achieved by the proposed joint pruning and caching strategy.
All the experiments utilize FP32 precision on an NVIDIA A6000 GPU.

\begin{figure*}[t]
  \centering
  \includegraphics[width=\textwidth]{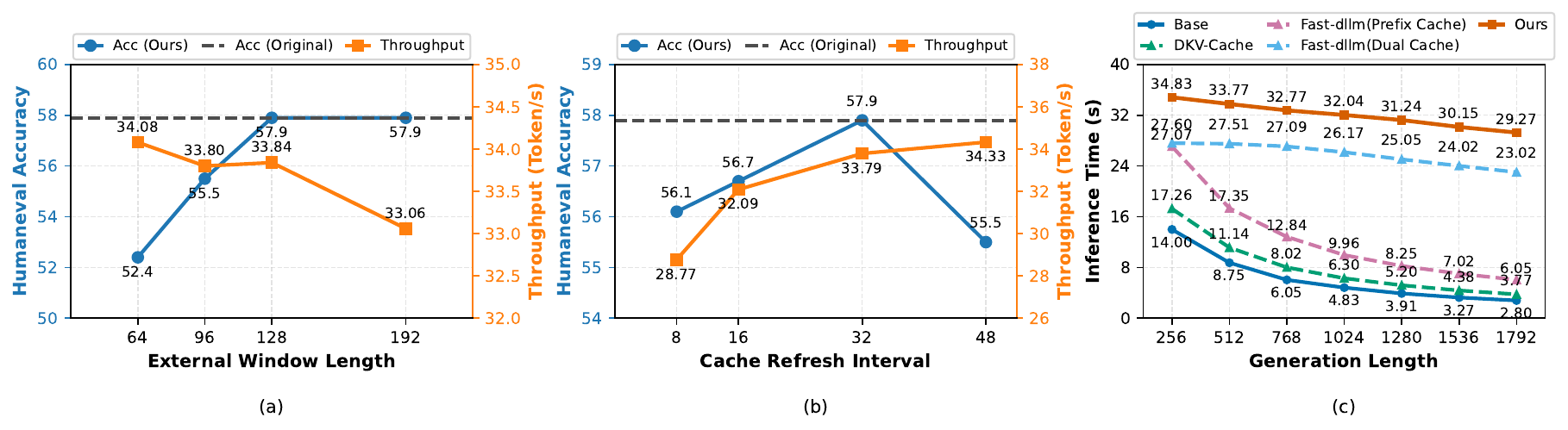}
  \caption{(a) \textbf{Ablation study on the external window length} for our proposed optimization strategy on Dream-Instruct  evaluated on HumanEval (0-shot), where the cache refresh cycle=32. (b) \textbf{Ablation study on the cache refresh interval} for our optimization strategy on Dream-Instruct evaluated on HumanEval (0-shot). (c) Inference time comparison across \textbf{generation lengths} for different acceleration methods on Dream-Instruct.}
  \label{fig:threepanels_wide}
\end{figure*}

\subsection{Main Results}

\paragraph{Window-based pruning incurs minimal performance degradation.}
Table~\ref{tab:token prunning} compares our pruning-only method (without caching) with Block Diffusion on the Dream model using $L$=16 and $L$=32. With a small window ($L$=16), Block Diffusion shows significant degradation, especially for Instruct, while Window-Diffusion preserves performance better. When $L$=32, Window-Diffusion nearly matches the original Dream model on the Base model, and performance recovers for Instruct as the window size increases. Overall, Window-Diffusion is more robust than Block Diffusion and better mitigates pruning-induced loss without the use of caching.

\paragraph{Performance and Efficiency Comparison with Existing Acceleration Methods.} Table~\ref{tab:perf_comparison} further summarizes the performance and inference efficiency of Window-Diffusion compared with several existing acceleration methods, including DKV-Cache \cite{ma2025dkvcachetc} and Fast-dLLM \cite{wu2025fastdllmta} (for fairness, parallel decoding is disabled). In addition to accuracy, we report the corresponding throughput (tokens/s) and speedup over the original model.

Overall, Window-Diffusion achieves consistent performance across all tasks while providing substantial inference acceleration. Our method maintains accuracy comparable to the original models on four benchmarks, while delivering significant speedups, with an average throughput improvement of approximately $5\times$.




    
    

\newcommand{\score}[1]{\textcolor{blue}{#1}}                 
\newcommand{\spd}[1]{\textcolor{yellow!60!black}{#1}}        
\newcommand{\cell}[3]{
  \begin{tabular}{@{}c@{}}
    \score{#1}\\
    #2 (\spd{#3})
  \end{tabular}
}

\begin{table}[!htbp]
  \centering
  \small
  \setlength{\tabcolsep}{4.2pt}
  \renewcommand{\arraystretch}{1.05}

  \begin{tabular}{@{} l c c c c @{}}
    \toprule
    \multirow{2}{*}{\textbf{Task}} &
    \multirow{2}{*}{\textbf{Length}} &
    \multirow{2}{*}{\textbf{Dream}} &
    \multicolumn{2}{c}{\textbf{Window-Diffusion}} \\
    \cmidrule(lr){4-5}
    & & & \textbf{WD-Static} & \textbf{WD-Adaptive} \\
    \midrule

    GSM8K & 256 &
    \scell{81.0}{17.8}{1.0$\times$} &
    \scell{82.9}{7.9}{2.3$\times$} &
    \scell{82.4}{3.5}{5.1$\times$} \\
    \midrule

    MATH & 512 &
    \scell{39.2}{58.8}{1.0$\times$} &
    \scell{38.5}{15.1}{3.9$\times$} &
    \scell{38.9}{6.4}{9.2$\times$} \\
    \midrule

    HumanEval & 768 &
    \scell{55.5}{125.9}{1.0$\times$} &
    \scell{58.5}{23.8}{5.3$\times$} &
    \scell{57.9}{2.9}{43.4$\times$} \\
    \midrule

    MBPP & 1024 &
    \scell{58.8}{217.8}{1.0$\times$} &
    \scell{55.4}{32.9}{6.6$\times$} &
    \scell{55.6}{2.2}{99.0$\times$} \\
    \bottomrule
  \end{tabular}

    \caption{
    Performance and inference efficiency of Dream-Instruct under fixed-length and adaptive-length inference across four tasks.
    Each cell reports task accuracy (top row) and inference latency in seconds with relative speedup to the Dream baseline (bottom row, {\color{thrblue}blue: latency} / {\color{sporange}orange: relative speedup}).
    }
  \label{tab:earlystop}
\end{table}

\paragraph{Adaptive-Length Inference Improves Efficiency with Minimal Accuracy Impact.}
Table~\ref{tab:earlystop} compares fixed-length (Dream, WD-Fixed) and adaptive-length (WD-Adaptive) inference on Dream-Instruct across four tasks.
Under fixed-length inference, generation always proceeds to a predefined maximum length, even after emitting \texttt{<eos>}, resulting in redundant computation.
In contrast, WD-Adaptive terminates decoding upon \texttt{<eos>}, allowing the generation length to adapt to individual instances.

As shown in Table~\ref{tab:earlystop}, adaptive-length inference consistently reduces latency across all tasks, with particularly large gains on code generation benchmarks such as HumanEval and MBPP.
Meanwhile, generation quality is largely preserved, with only minor variations compared to fixed-length inference.
These results indicate that adaptive-length inference is an effective complementary mechanism for improving inference efficiency in diffusion language models.

\subsection{Ablations and Analysis}

\paragraph{Effect of External Window Length.}
Figure~\ref{fig:threepanels_wide}(a) analyzes the impact of the external window length on model accuracy and inference throughput. We fix the cache refresh cycle to 32, set the internal window length to 16, and use a generation length of 512. Experiments are conducted on the HumanEval dataset in the 0-shot setting using Dream-Base. The results show that model performance improves as the window length increases and gradually saturates, confirming the rapidly diminishing marginal contribution of masked tokens. Meanwhile, as the window length grows, inference throughput decreases moderately due to the increased amount of context to be processed, while the overall slowdown remains limited.

Balancing inference efficiency and model performance, we ultimately set the external window length to 128.

\paragraph{Effect of Cache Refresh Cycle.} 
Figure~\ref{fig:threepanels_wide}(b) examines the effect of the cache refresh cycle on the Base model. We set the internal window length to 16, the external window length to 128, and the generation length to 512, and evaluate Dream-Base on the HumanEval dataset in the 0-shot setting.

The results show that inference throughput steadily increases as the refresh cycle grows, but the improvement gradually plateaus. Meanwhile, model accuracy exhibits a non-monotonic trend, first improving and then degrading. This behavior stems from our implementation strategy: decoded tokens within the internal window are not immediately written to the KV cache, but are recomputed in full until their KV representations are updated at the next cache refresh. As the refresh cycle increases, more tokens within the internal window require full computation, which offsets part of the throughput gains brought by less frequent cache updates. Conversely, when the refresh cycle is small, newly decoded tokens are cached rapidly while their KV representations are still unstable, leading to a drop in model accuracy.

Balancing inference efficiency and model performance, we set the cache refresh cycle to 32.

\paragraph{Effect of Generation Length.}
Figure~\ref{fig:threepanels_wide}(c) reports the inference time under different generation lengths for various acceleration methods. We fix a single input instance and progressively increase the generation length, comparing the original model, DKV-Cache, Fast-dLLM, and Window-Diffusion.

As the generation length increases, the inference cost becomes increasingly sensitive to sequence length: the overall computation grows rapidly in long-generation regimes, and the average compute cost per token also increases as attention and context processing become more expensive with longer sequences. The methods exhibit clearly different behaviors. The original model is constrained by its high time complexity and incurs substantially higher cost for long sequences. DKV-Cache and Fast-dLLM (Prefix-Cache) only cache decoded tokens and do not reduce the computation on masked tokens, leading to limited improvement in inference efficiency. Fast-dLLM (Dual-Cache) further caches undecoded tokens and reduces repeated computation to some extent; however, periodic KV-cache refresh still makes it difficult to avoid full-sequence involvement during inference.

In contrast, Window-Diffusion consistently maintains a strong speedup across different generation lengths, and its relative advantage persists as the generation length increases. This is because Window-Diffusion shortens the masked-token sequence via token-level pruning, thereby reducing the overall computation for long-form generation.

\section{Conclusion}

In this work, we conduct a token-level analysis of diffusion language model inference and identify substantial computational redundancy arising from spatial locality, rapidly saturating context dependence, and short-term representational stability. Motivated by these observations, we propose \textbf{Window-Diffusion}, a training-free inference framework that combines a dual-window token organization with phase-level KV reuse to reduce unnecessary computation while preserving generation quality. Extensive experiments across multiple tasks and model settings demonstrate that our method achieves consistent and significant inference speedups with minimal performance degradation, highlighting its practical effectiveness for efficient diffusion-based text generation.


\bibliographystyle{named}
\bibliography{ijcai26}

\begin{thebibliography}{}

\bibitem[\protect\citeauthoryear{Arriola \bgroup \em et al.\egroup
  }{2025}]{arriola2025blockdi}
Marianne Arriola, Aaron Gokaslan, Justin~T Chiu, Zhihan Yang, Zhixuan Qi, Jiaqi
  Han, Subham~Sekhar Sahoo, and Volodymyr Kuleshov.
\newblock Block diffusion: Interpolating between autoregressive and diffusion
  language models.
\newblock {\em ArXiv}, abs/2503.09573, 2025.

\bibitem[\protect\citeauthoryear{Austin \bgroup \em et al.\egroup
  }{2021a}]{Austin2021StructuredDD}
Jacob Austin, Daniel~D. Johnson, Jonathan Ho, Daniel Tarlow, and Rianne van~den
  Berg.
\newblock Structured denoising diffusion models in discrete state-spaces.
\newblock {\em ArXiv}, abs/2107.03006, 2021.

\bibitem[\protect\citeauthoryear{Austin \bgroup \em et al.\egroup
  }{2021b}]{austin2021programsw}
Jacob Austin, Augustus Odena, Maxwell Nye, Maarten Bosma, Henryk Michalewski,
  David Dohan, Ellen Jiang, Carrie~J. Cai, Michael Terry, Quoc~V. Le, and
  Charles Sutton.
\newblock Program synthesis with large language models.
\newblock {\em ArXiv}, abs/2108.07732, 2021.

\bibitem[\protect\citeauthoryear{Chen \bgroup \em et al.\egroup
  }{2021}]{chen2021evaluatingll}
Mark Chen, Jerry Tworek, Heewoo Jun, Qiming Yuan, Henrique Pond{\'e}, Jared
  Kaplan, Harrison Edwards, Yura Burda, Nicholas Joseph, Greg Brockman, Alex
  Ray, Raul Puri, Gretchen Krueger, Michael Petrov, Heidy Khlaaf, Girish
  Sastry, Pamela Mishkin, Brooke Chan, Scott Gray, Nick Ryder, Mikhail Pavlov,
  Alethea Power, Lukasz Kaiser, Mo~Bavarian, Clemens Winter, Phil Tillet,
  Felipe~Petroski Such, David~W. Cummings, Matthias Plappert, Fotios Chantzis,
  Elizabeth Barnes, Ariel Herbert-Voss, William~H. Guss, Alex Nichol, Igor
  Babuschkin, Suchir Balaji, Shantanu Jain, Andrew Carr, Jan Leike, Josh
  Achiam, Vedant Misra, Evan Morikawa, Alec Radford, Matthew~M. Knight, Miles
  Brundage, Mira Murati, Katie Mayer, Peter Welinder, Bob McGrew, Dario Amodei,
  Sam McCandlish, Ilya Sutskever, and Wojciech Zaremba.
\newblock Evaluating large language models trained on code.
\newblock {\em ArXiv}, abs/2107.03374, 2021.

\bibitem[\protect\citeauthoryear{Cobbe \bgroup \em et al.\egroup
  }{2021}]{cobbe2021trainingvt}
Karl Cobbe, Vineet Kosaraju, Mo~Bavarian, Mark Chen, Heewoo Jun, Lukasz Kaiser,
  Matthias Plappert, Jerry Tworek, Jacob Hilton, Reiichiro Nakano, Christopher
  Hesse, and John Schulman.
\newblock Training verifiers to solve math word problems.
\newblock {\em ArXiv}, abs/2110.14168, 2021.

\bibitem[\protect\citeauthoryear{Feng \bgroup \em et al.\egroup
  }{2025}]{feng2025theoreticalba}
Guhao Feng, Yihan Geng, Jian Guan, Wei Wu, Liwei Wang, and Di~He.
\newblock Theoretical benefit and limitation of diffusion language model.
\newblock {\em ArXiv}, abs/2502.09622, 2025.

\bibitem[\protect\citeauthoryear{Hendrycks \bgroup \em et al.\egroup
  }{2021}]{hendrycks2021measuringmp}
Dan Hendrycks, Collin Burns, Saurav Kadavath, Akul Arora, Steven Basart, Eric
  Tang, Dawn~Xiaodong Song, and Jacob Steinhardt.
\newblock Measuring mathematical problem solving with the math dataset.
\newblock {\em ArXiv}, abs/2103.03874, 2021.

\bibitem[\protect\citeauthoryear{Li \bgroup \em et al.\egroup
  }{2025}]{li2025aso}
Tianyi Li, Mingda Chen, Bowei Guo, and Zhiqiang Shen.
\newblock A survey on diffusion language models.
\newblock {\em ArXiv}, abs/2508.10875, 2025.

\bibitem[\protect\citeauthoryear{Lin \bgroup \em et al.\egroup
  }{2025}]{lin2025quantizationmd}
Haokun Lin, Haobo Xu, Yichen Wu, Ziyu Guo, Renrui Zhang, Zhichao Lu, Ying Wei,
  Qingfu Zhang, and Zhenan Sun.
\newblock Quantization meets dllms: A systematic study of post-training
  quantization for diffusion llms.
\newblock {\em ArXiv}, abs/2508.14896, 2025.

\bibitem[\protect\citeauthoryear{Liu \bgroup \em et al.\egroup
  }{2025a}]{liu2025sequentialdl}
Yangzhou Liu, Yue Cao, Hao Li, Gen Luo, Zhe Chen, Weiyun Wang, Xiaobo Liang,
  Biqing Qi, Lijun Wu, Changyao Tian, Yanting Zhang, Yuqiang Li, Tong Lu,
  Yu~Qiao, Jifeng Dai, and Wenhai Wang.
\newblock Sequential diffusion language models.
\newblock {\em ArXiv}, abs/2509.24007, 2025.

\bibitem[\protect\citeauthoryear{Liu \bgroup \em et al.\egroup
  }{2025b}]{liu2025dllmcachead}
Zhiyuan Liu, Yicun Yang, Yaojie Zhang, Junjie Chen, Chang Zou, Qingyuan Wei,
  Shaobo Wang, and Linfeng Zhang.
\newblock dllm-cache: Accelerating diffusion large language models with
  adaptive caching.
\newblock {\em ArXiv}, abs/2506.06295, 2025.

\bibitem[\protect\citeauthoryear{Lou \bgroup \em et al.\egroup
  }{2023}]{lou2023discretedm}
Aaron Lou, Chenlin Meng, and Stefano Ermon.
\newblock Discrete diffusion modeling by estimating the ratios of the data
  distribution.
\newblock In {\em International Conference on Machine Learning}, 2023.

\bibitem[\protect\citeauthoryear{Ma \bgroup \em et al.\egroup
  }{2025}]{ma2025dkvcachetc}
Xinyin Ma, Runpeng Yu, Gongfan Fang, and Xinchao Wang.
\newblock dkv-cache: The cache for diffusion language models.
\newblock {\em ArXiv}, abs/2505.15781, 2025.

\bibitem[\protect\citeauthoryear{Nie \bgroup \em et al.\egroup
  }{2024}]{nie2024scalingum}
Shen Nie, Fengqi Zhu, Chao Du, Tianyu Pang, Qian Liu, Guangtao Zeng, Min Lin,
  and Chongxuan Li.
\newblock Scaling up masked diffusion models on text.
\newblock {\em ArXiv}, abs/2410.18514, 2024.

\bibitem[\protect\citeauthoryear{Nie \bgroup \em et al.\egroup
  }{2025}]{nie2025largeld}
Shen Nie, Fengqi Zhu, Zebin You, Xiaolu Zhang, Jingyang Ou, Jun Hu, Jun Zhou,
  Yankai Lin, Jirong Wen, and Chongxuan Li.
\newblock Large language diffusion models.
\newblock {\em ArXiv}, abs/2502.09992, 2025.

\bibitem[\protect\citeauthoryear{Peng \bgroup \em et al.\egroup
  }{2025}]{peng2025howea}
Han Peng, Peiyu Liu, Zican Dong, Daixuan Cheng, Junyi Li, Yiru Tang, Shuo Wang,
  and Wayne~Xin Zhao.
\newblock How efficient are diffusion language models? a critical examination
  of efficiency evaluation practices.
\newblock {\em ArXiv}, abs/2510.18480, 2025.

\bibitem[\protect\citeauthoryear{Sahoo \bgroup \em et al.\egroup
  }{2024}]{sahoo2024simpleae}
Subham~Sekhar Sahoo, Marianne Arriola, Yair Schiff, Aaron Gokaslan, Edgar
  Marroquin, Justin~T Chiu, Alexander Rush, and Volodymyr Kuleshov.
\newblock Simple and effective masked diffusion language models.
\newblock {\em ArXiv}, abs/2406.07524, 2024.

\bibitem[\protect\citeauthoryear{Wang \bgroup \em et al.\egroup
  }{2025}]{wang2025diffusionlc}
Xu~Wang, Chenkai Xu, Yijie Jin, Jiachun Jin, Hao Zhang, and Zhijie Deng.
\newblock Diffusion llms can do faster-than-ar inference via discrete diffusion
  forcing.
\newblock {\em ArXiv}, abs/2508.09192, 2025.

\bibitem[\protect\citeauthoryear{Wu \bgroup \em et al.\egroup
  }{2025}]{wu2025fastdllmta}
Chengyue Wu, Hao Zhang, Shuchen Xue, Zhijian Liu, Shizhe Diao, Ligeng Zhu, Ping
  Luo, Song Han, and Enze Xie.
\newblock Fast-dllm: Training-free acceleration of diffusion llm by enabling kv
  cache and parallel decoding.
\newblock {\em ArXiv}, abs/2505.22618, 2025.

\bibitem[\protect\citeauthoryear{Xu and Yang}{2025}]{xu2025dllmquantqd}
Chen Xu and Dawei Yang.
\newblock Dllmquant: Quantizing diffusion-based large language models.
\newblock {\em ArXiv}, abs/2508.14090, 2025.

\bibitem[\protect\citeauthoryear{Ye \bgroup \em et al.\egroup
  }{2025}]{ye2025dream7d}
Jiacheng Ye, Zhihui Xie, Lin Zheng, Jiahui Gao, Zirui Wu, Xin Jiang, Zhenguo
  Li, and Lingpeng Kong.
\newblock Dream 7b: Diffusion large language models.
\newblock {\em ArXiv}, abs/2508.15487, 2025.

\bibitem[\protect\citeauthoryear{Zhang \bgroup \em et al.\egroup
  }{2025a}]{zhang2025diffusion}
Siyue Zhang, Yilun Zhao, Liyuan Geng, Arman Cohan, Luu~Anh Tuan, and Chen Zhao.
\newblock Diffusion vs. autoregressive language models: A text embedding
  perspective.
\newblock In {\em Proceedings of the 2025 Conference on Empirical Methods in
  Natural Language Processing}, pages 4273--4303, 2025.

\bibitem[\protect\citeauthoryear{Zhang \bgroup \em et al.\egroup
  }{2025b}]{zhang2025quantdllmpe}
Tianao Zhang, Zhiteng Li, Xianglong Yan, Haotong Qin, Yong Guo, and Yulun
  Zhang.
\newblock Quant-dllm: Post-training extreme low-bit quantization for diffusion
  large language models.
\newblock {\em ArXiv}, abs/2510.03274, 2025.

\bibitem[\protect\citeauthoryear{Zhu \bgroup \em et al.\egroup
  }{2025}]{zhu2025llada1v}
Fengqi Zhu, Rongzheng Wang, Shen Nie, Xiaolu Zhang, Chunwei Wu, Jun Hu, Jun
  Zhou, Jianfei Chen, Yankai Lin, Jirong Wen, and Chongxuan Li.
\newblock Llada 1.5: Variance-reduced preference optimization for large
  language diffusion models.
\newblock {\em ArXiv}, abs/2505.19223, 2025.

\end{thebibliography}

\newpage

\appendix

\section{Appendix}

\subsection{Experimental Settings for Main Results on Dream}
\label{appendix:hyperparameters}

This appendix summarizes the hyperparameter configurations used in the main Dream experiments reported in Table~2.
For the number of in-context examples (\#shots) and the maximum generation length, we follow the official settings provided in the Dream codebase to ensure consistency with the original results.

To align the comparison with prior work, we configure DKV-Cache with a cache refresh interval of 4, as suggested in its original implementation.
For Fast-dLLM, we set the block size to 32, enable PrefixCache, and disable parallel decoding to ensure a fair comparison under comparable decoding conditions.

For Window-Diffusion, we set the external window length to 128, the internal window length to 16, and the cache refresh interval to 32 across all experiments.
Unless otherwise specified, early stopping is disabled in the main experiments.

The task-specific hyperparameters for Dream models are summarized in Table~\ref{tab:hparams_by_task_dream}.


\begin{table}[!htbp]
  \centering
  \resizebox{\columnwidth}{!}{%
    \begin{tabular}{@{} l l c c @{}}
      \toprule
      \textbf{Model} & \textbf{Task} & \textbf{\#Shots} & \textbf{Max generation length} \\
      \midrule
      \multirow{4}{*}{Base}
        & GSM8K      & 8 & 256 \\
        & MATH       & 4 & 512 \\
        & HumanEval  & 0 & 512 \\
        & MBPP       & 3 & 512 \\
      \midrule
      \multirow{4}{*}{Instruct}
        & GSM8K      & 0 & 256 \\
        & MATH       & 0 & 512 \\
        & HumanEval  & 0 & 768 \\
        & MBPP       & 0 & 1024 \\
      \bottomrule
    \end{tabular}%
  }
  \caption{Hyperparameters used for each Dream model and task.}
  \label{tab:hparams_by_task_dream}
\end{table}

\subsection{Additional Results on LLaDA-Base Model}
\label{appendix:llada}

To further validate the generality of our method, we report additional experimental results of Window-Diffusion on the LLaDA-Base model in this appendix. The overall experimental pipeline and evaluation benchmarks are consistent with those used in the main Dream experiments, while task-specific hyperparameters follow standard LLaDA evaluation practices.

To ensure fair comparisons, we adopt acceleration method configurations commonly used in prior work on LLaDA. Notably, DKV-Cache was originally evaluated only on the LLaDA-Instruct model. In this work, we reimplement DKV-Cache on LLaDA-Base by directly applying its original Instruct configuration, keeping all cache-related hyperparameters unchanged, with the cache refresh interval set to 8 and the block size set to 32. For Fast-dLLM, we report results for both the Prefix-Cache and Dual-Cache variants; in both cases, the block size is fixed to 32 and parallel decoding is disabled to ensure comparable decoding conditions across methods.

For Window-Diffusion on LLaDA-Base, we use an external window length of 64 and fix the internal window length to 16 across all tasks. The cache refresh interval is set to 32, and early stopping is disabled. Task-specific hyperparameters are summarized in Table~\ref{tab:hparams_by_task_llada}.

Under these settings, the performance comparison of different acceleration methods on LLaDA-Base is reported in Table~\ref{tab:llada_base_perf}. We observe that while the Dual-Cache variant of Fast-dLLM further improves inference throughput on some tasks, it also leads to more pronounced degradation in generation quality. In contrast, Window-Diffusion consistently achieves the \emph{highest inference speedups} among all compared methods, while maintaining competitive generation quality across tasks. These results indicate that Window-Diffusion offers an effective and stable acceleration strategy under an explicit speed--quality trade-off, and further demonstrate its robustness across different diffusion language models.

\begin{table}[t]
  \centering
  \resizebox{\columnwidth}{!}{%
  \begin{tabular}{@{} l l c c @{}}
    \toprule
    \textbf{Model} & \textbf{Task} & \textbf{\#Shots} & \textbf{Max generation length} \\
    \midrule
    \multirow{4}{*}{Base}
      & GSM8K     & 4 & 256 \\
      & MATH      & 4 & 256 \\
      & HumanEval & 0 & 512 \\
      & MBPP      & 3 & 512 \\
    \midrule
  \end{tabular}}
  \caption{Hyperparameters used for the LLaDA-Base model and tasks.}
  \label{tab:hparams_by_task_llada}
\end{table}

\begin{table}
\centering
\small
\setlength{\tabcolsep}{6pt}
\renewcommand{\arraystretch}{1.15}

\begin{tabular}{>{\centering\arraybackslash}m{1.05cm} cccc}
\toprule
& \multicolumn{4}{c}{\textbf{Base}} \\
\cmidrule(lr){2-5}
\textbf{Method} &
\textbf{GSM8K} & \textbf{Math} & \textbf{HumanEval} & \textbf{MBPP} \\
\midrule

LLaDA
& \scell{70.7}{4.49}{1.0$\times$}
& \scell{30.1}{5.06}{1.0$\times$}
& \scell{32.9}{6.59}{1.0$\times$}
& \scell{39.2}{4.18}{1.0$\times$} \\
\midrule

DKV-Cache
& \scell{68.7}{8.07}{1.80$\times$}
& \scell{27.8}{8.79}{1.74$\times$}
& \scell{31.7}{9.08}{1.38$\times$}
& \scell{38.2}{6.61}{1.58$\times$} \\
\midrule

Fast-dLLM
(Prefix-Cache)
& \scell{68.8}{14.71}{3.28$\times$}
& \scell{28.5}{15.39}{3.04$\times$}
& \scell{30.5}{12.03}{1.83$\times$}
& \scell{37.8}{10.37}{2.48$\times$} \\
\midrule

Fast-dLLM
(Dual-Cache)
& \scell{67.9}{17.28}{3.85$\times$}
& \scell{25.5}{18.37}{3.63$\times$}
& \scell{27.4}{20.25}{3.07$\times$}
& \scell{36.0}{16.71}{4.00$\times$} \\
\midrule

\textbf{Window-Diffusion}
& \scell{68.5}{25.14}{5.60$\times$}
& \scell{26.2}{25.93}{5.12$\times$}
& \scell{28.0}{27.91}{4.24$\times$}
& \scell{38.2}{24.63}{5.90$\times$} \\
\bottomrule
\end{tabular}

\caption{Performance comparison of acceleration methods on the LLaDA Base model.
Each cell reports accuracy (top row) and decoding throughput in tokens per second (middle row, {\color{thrblue}blue: tokens per second}) and relative speedup to the LLaDA baseline (bottom row, {\color{sporange}orange: relative speedup}).
Window-Diffusion uses an internal window size of $L=16$ with a refresh cycle of 32, and early stopping is disabled.
}
\label{tab:llada_base_perf}
\end{table}

\end{document}